\documentclass[letterpaper]{article} 
\usepackage[draft]{aaai2027}  
\usepackage[hyphens]{url}  
\usepackage{graphicx} 
\usepackage{natbib}  
\usepackage{caption} 
\usepackage{algorithm}
\usepackage{algorithmic}

\usepackage{newfloat}
\usepackage{listings}
\DeclareCaptionStyle{ruled}{labelfont=normalfont,labelsep=colon,strut=off} 
\floatstyle{ruled}
\newfloat{listing}{tb}{lst}{}
\floatname{listing}{Listing}

\usepackage{booktabs}

\usepackage{multirow}
\usepackage{pifont}     
\usepackage{subcaption} 
\usepackage{graphicx}   
\usepackage[table]{xcolor}
\usepackage{amsmath}
\usepackage{amssymb}
\usepackage{xcolor}
\usepackage{cleveref}
\usepackage{adjustbox}

\title{Towards Robust Vision-Language Reasoning: A Multi-Stage Optimization Pipeline with Masked Diffusion Model}
\author{
    Shunming Liu\textsuperscript{\rm 1,\rm 2},
    Pei Huang\textsuperscript{\rm 3},
    Yajing Sun\textsuperscript{\rm 2}\corresponding,
    Huaxiao Liu\textsuperscript{\rm 1}\corresponding
}
\affiliations{
    \textsuperscript{\rm 1}College of Computer Science and Technology, Jilin University, Changchun, China\\
    \textsuperscript{\rm 2}Yinwang Intelligent Technology Co., Ltd., Huawei, China\\
    \textsuperscript{\rm 3}Stanford University, USA\\
    smliu23@mails.jlu.edu.cn, liuhuaxiao@jlu.edu.cn
}

\begin{document}

\maketitle

\begin{abstract}
Discrete diffusion generation is emerging as a competitive alternative to autoregressive generation by denoising the entire token sequence through iterative refinement. Its global planning and progressive correction capabilities are particularly well suited to autonomous driving. However, transferring this generation paradigm from general-purpose settings to driving introduces substantial domain gaps in both training and inference. To unlock its potential, we propose a multi-stage optimization pipeline for standard vision-language model architectures, without auxiliary action experts or specialized mixture-of-experts Transformer, spanning post-training and inference. During supervised fine-tuning, we optimize the representation of planning inputs to suppress hallucinations in task responses. During reinforcement learning, we introduce metric-guided diversity filtering to identify stronger learning starting points and improve policy exploration. At inference time, we propose Dynamic Autonomous Driving Decoding (DAD), which adaptively adjusts decoding modes according to scene complexity. Based on this pipeline, we develop Masked Diffusion Drive (MD-Drive), improving its performance by 29.7\% over the unoptimized model. Experiments on the NAVSIM closed loop benchmark and the nuScenes and DriveCoT open loop datasets show that MD-Drive outperforms autoregressive VLM baselines with the same architecture, while achieving a more favorable balance between performance and computational cost than systems with different architectures through discrete diffusion training and inference.

\end{abstract}

\section{Introduction}
Vision-Language Models (VLMs) have made breakthrough progress in integrating visual representations with natural language semantics, significantly advancing complex applications ranging from multimodal dialogue to autonomous driving~\cite{openai2024gpt4technicalreport,nvidia2026alpamayor1bridgingreasoningaction}. Although current success is primarily built upon the Autoregressive (AR) paradigm, the inherent limitations of this paradigm cannot be ignored. Since autoregressive models are trained based on\begin{figure}[ht]
    \centering
    \includegraphics[scale=0.28]{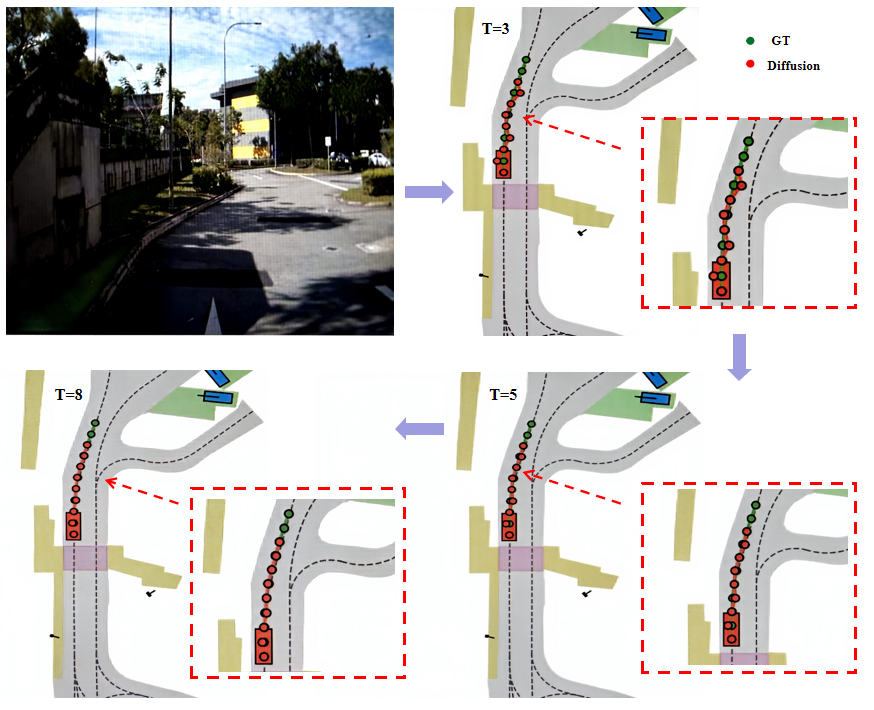}
    \caption{Illustration of discrete diffusion generation for autonomous driving. Starting from a coarse trajectory, the model iteratively refines the full token sequence into an increasingly accurate 4-second prediction.
}
    \label{fig:introduction}
\end{figure} next-token prediction, their inference process relies on sequential generation. This mechanism easily leads to error accumulation, which in turn restricts the model's deep reasoning and long-term planning capabilities in complex scenarios~\cite{The_Pitfalls_of_Next-Token_Prediction,Computational-Statistical-Tradeoffs-at-the-Next-Token-Prediction}. 

Recently, the rise of discrete diffusion Large Language Models (dLLMs) provides an attractive alternative to address the aforementioned challenges. For example, models such as Mercury~\cite{labs2025mercuryultrafastlanguagemodels} and Gemini~\cite{geminidiffusion} demonstrate superior performance over autoregressive models (AR LLMs) in code generation tasks and exhibit several core advantages~\cite{gong2025scalingdiffusionlanguagemodels,gong2025diffucoderunderstandingimprovingmasked,Large-Language-Diffusion-Models,d1}: \ding{172} Their inference is more flexible, supporting parallel decoding to generate multiple tokens simultaneously; \ding{173} They possess stronger global planning capabilities, coordinating bidirectional contextual information for collaborative reasoning during the generation process to ensure high reliability of the output results. As illustrated in Fig.~\ref{fig:introduction}, the trajectory iteratively optimizes over the entire sequence as the time steps change, evolving from a rough to a smooth trajectory. These characteristics of the discrete diffusion paradigm not only effectively compensate for the inherent flaws of autoregressive models but also more precisely meet the requirements of autonomous driving tasks for robust decision-making and global coordination in complex dynamic environments.

However, several critical issues arise when employing the existing training and inference mechanisms of discrete diffusion. First, there is a significant lack of structural representation. During planning tasks, the model frequently generates redundant tokens and unstructured noise, leading to a mismatch between diffusion outputs and the specific information required for driving. Treating continuous trajectory waypoints as pure text sequences introduces noise that undermines the robustness of downstream trajectory execution. Second, the model faces learning efficiency and convergence bottlenecks. During the reinforcement learning stage, it fails to capture key information in hard scenarios, which causes performance gains to stall or even results in policy degradation. Third, there is a mismatch between inference logic and driving scenarios. Traditional confidence-based greedy decoding strategies violate driving logic and ignore domain-specific priorities, thereby introducing significant uncertainty into autonomous driving systems.

To address these issues, we propose a \textbf{multi-stage optimization pipeline} that achieves system-level optimization across three dimensions. In the supervised learning stage, we introduce a trajectory discretization strategy. By reconstructing trajectory representations, this approach eliminates noise caused by text serialization and enhances structural constraints on model outputs. In the reinforcement learning stage, a diversity filtering mechanism based on pass-rate metrics is proposed to identify and select the most valuable learning starting points, thereby breaking through performance bottlenecks and maximizing policy improvement efficiency. Finally, in the inference stage, we pioneer a Dynamic Autonomous Driving System Decoding strategy. This strategy utilizes confidence trends as the criterion for judging driving risk and deconstructs inference logic into two anthropomorphic thinking modes, from far to near and from near to far, achieving adaptive decision-making across scenarios of varying complexity.

Evaluation on NAVSIM, nuScenes, and DriveCoT shows that MD-Drive outperforms autoregressive standard Vision-Language-Model baselines. Compared with models using different architectures and generation methods, it achieves competitive performance with the smallest training set. Ablation studies on NAVSIM further show that the proposed SFT, RL, and inference optimizations each contribute to performance, together yielding a 29.7\% improvement. Our main contributions are summarized as follows:
\begin{itemize}
\item We propose a synergistic \textit{Multi-stage Optimization Pipeline} for discrete diffusion token generation within standard VLM architectures. We introduce corresponding optimization methods in the Supervised Fine-Tuning (SFT), Group Relative Policy Optimization (GRPO), and inference stages to improve overall model performance.
\item We introduce Dynamic Autonomous Driving Decoding (DAD), an adaptive inference strategy that automatically selects an appropriate decoding order according to scene complexity.
\item We develop \textit{MD-Drive}, an autonomous driving system built upon the proposed optimization pipeline. Extensive experiments on NAVSIM, nuScenes, and DriveCoT demonstrate that the pipeline consistently improves the performance of discrete diffusion generation.
\end{itemize}

\section{Related Work}
\subsection{Autonomous Driving Models}
Recent autonomous driving models differ in both architecture and generation paradigm. DrivingGPT~\cite{chen2024drivinggptunifyingdrivingworld} and FSDrive~\cite{zeng2025futuresightdrive} adopt unified VLM architectures without additional action experts or MoE modules, using autoregressive generation for visual and trajectory tokens. ReCogDrive~\cite{li2025recogdrive} combines an autoregressive VLM with an extra continuous diffusion planner for trajectory generation, while Epona~\cite{zhang2025epona} adopts a modular framework with autoregressive temporal generation and within-frame continuous diffusion. ViLaD~\cite{cui2025viladlargevisionlanguage} replaces autoregressive decoding in a standard VLM with masked discrete diffusion for parallel driving-sequence generation and iterative refinement. DriveVLA-W0~\cite{li2025drivevlaW0} augments the VLM backbone with a lightweight MoE action expert and supports both autoregressive and continuous flow-matching action generation. WAM-Flow~\cite{xu2025wamFLOW} retains a single VLM backbone but employs discrete flow matching for parallel trajectory-token refinement, whereas WAM-Diff~\cite{xu2025wamDIFF} introduces an MoE architecture with masked discrete diffusion. MVLAD-AD~\cite{zhang2026efficientexplainableendtoendautonomous} uses a unified VLM without an independent action expert or MoE module and generates discretized action tokens and language outputs through masked discrete diffusion.

\subsection{Discrete Diffusion Vision-Language Model}
Autoregressive vision-language models have achieved strong performance in multimodal understanding, reasoning, and instruction following by aligning pretrained vision encoders with large language models~\cite{wang2024qwen2vlenhancingvisionlanguagemodels,bai2025qwen25vltechnicalreport,yang2025qwen3technicalreport,guo-etal-2025-mammoth}. Extending diffusion language models to vision-language tasks, however, remains an emerging direction. LaViDa~\cite{li2025lavida} demonstrates that discrete diffusion models can achieve efficient multimodal modeling through non-causal decoding and bidirectional attention. LLaDA-V~\cite{you2025lladavlargelanguagediffusion} further introduces instruction tuning into large-scale discrete diffusion models, attaining performance comparable to autoregressive counterparts while improving long-form generation and reasoning efficiency. MMaDA~\cite{MMaDA} unifies visual understanding and image generation within a masked discrete diffusion Transformer, whereas Dimple~\cite{yu2025dimplediscretediffusionmultimodal} improves training stability through an autoregressive-then-diffusion paradigm. LaViDa-O~\cite{li2025lavidaoelasticlargemasked} extends masked diffusion to high-precision understanding and high-resolution generation using Elastic-MoT and planning-introspection mechanisms. DreamVL~\cite{ye2026dreamvldreamvlaopen} further applies the diffusion paradigm to robotic manipulation, unifying visual perception, reasoning, and action planning.

\begin{figure*}[ht]
\centering
\includegraphics[width=0.85\textwidth]{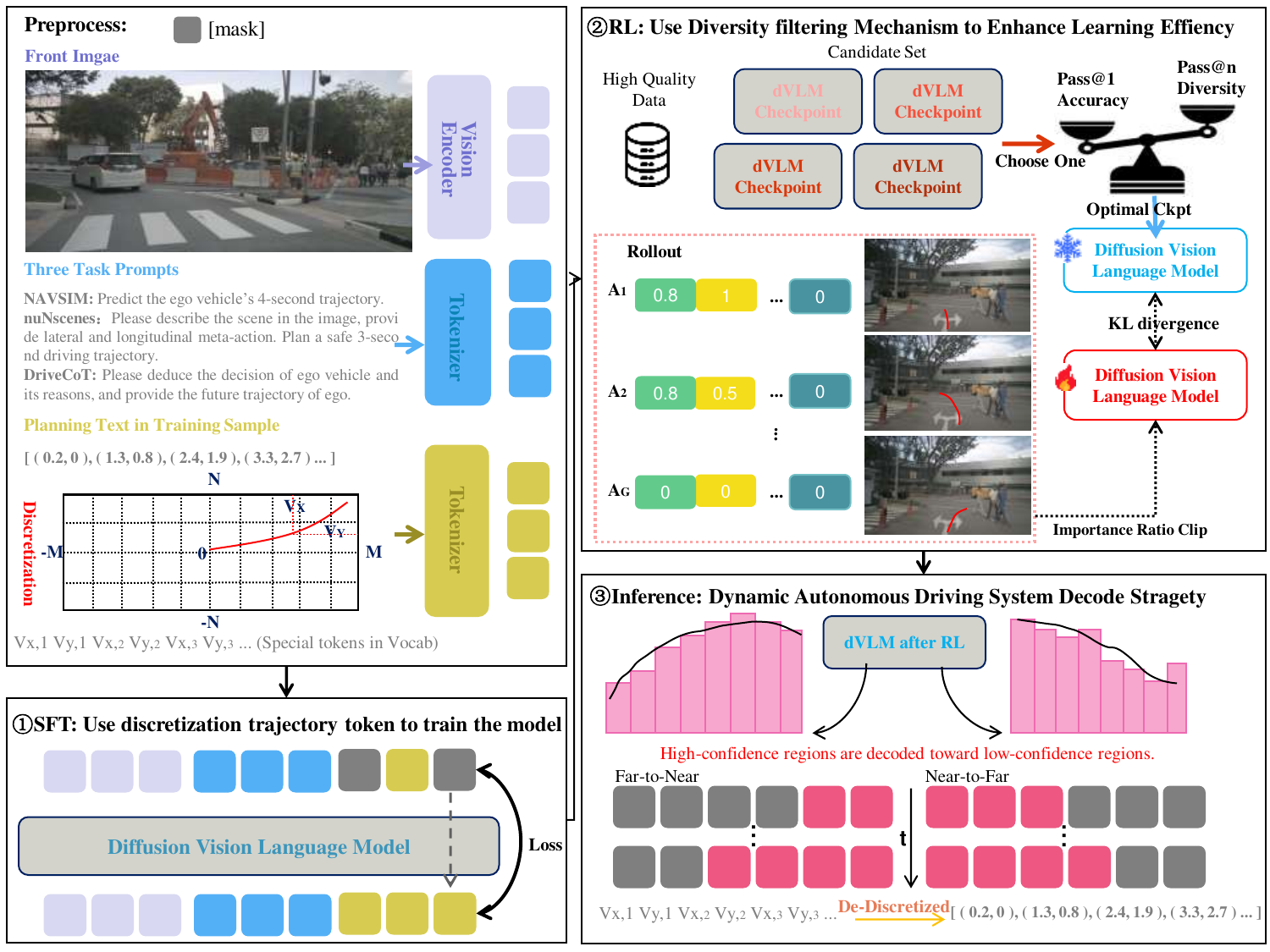}
\caption{Overview of Multi-Stage Optimization Pipeline with Masked Diffusion Model. Supervised Fine-Tuning: Trajectory discretization tokens are pre-stored in the vocabulary. During the processing of training samples, the planning components are discretized and replaced with corresponding special tokens to facilitate supervised learning. Reinforcement Learning: Checkpoints of varying maturity levels are screened based on diversity and accuracy metrics, after which they are fed into the Group Relative Policy Optimization (GRPO) training process. Inference: dVLM exhibits distinct confidence trends across different driving scenarios. DAD mechanism transforms these trends into specialized inference sequences to optimize output quality.}
\label{fig:overview}
\end{figure*}
\section{Method}
As shown in Fig.~\ref{fig:overview}, we propose a multi-stage optimization pipeline covering three stages. First, we introduce trajectory discretization, an optimization scheme for the supervised learning stage designed to simplify the structural representation of inputs. Second, we present a diversity filtering mechanism guided by key metrics, which serves as a reinforcement learning optimization solution to resolve efficiency bottlenecks and enhance model performance. Finally, we introduce a new decoding strategy DAD for the inference stage, which maps the trend of the model's output representations for a scene into two anthropomorphic autonomous driving reasoning modes.
\subsection{SFT: Trajectory Discretization}
In other continuous domains such as image generation~\cite{Zero-Shot-Text-to-Image-Generation,Scaling-Autoregressive-Models-for-Content-Rich-Text-to-Image-Generation} and audio generation~\cite{oord2016wavenetgenerativemodelraw}, discretizing the output space and predicting categorical distributions for each step has proven to be an effective method. To eliminate the noise introduced by continuous trajectory coordinate sequences in the planning phase and enhance the robustness\begin{figure*}[htbp]
    \centering
    \begin{subfigure}[t]{0.24\textwidth}
        \centering
        \includegraphics[width=\linewidth]{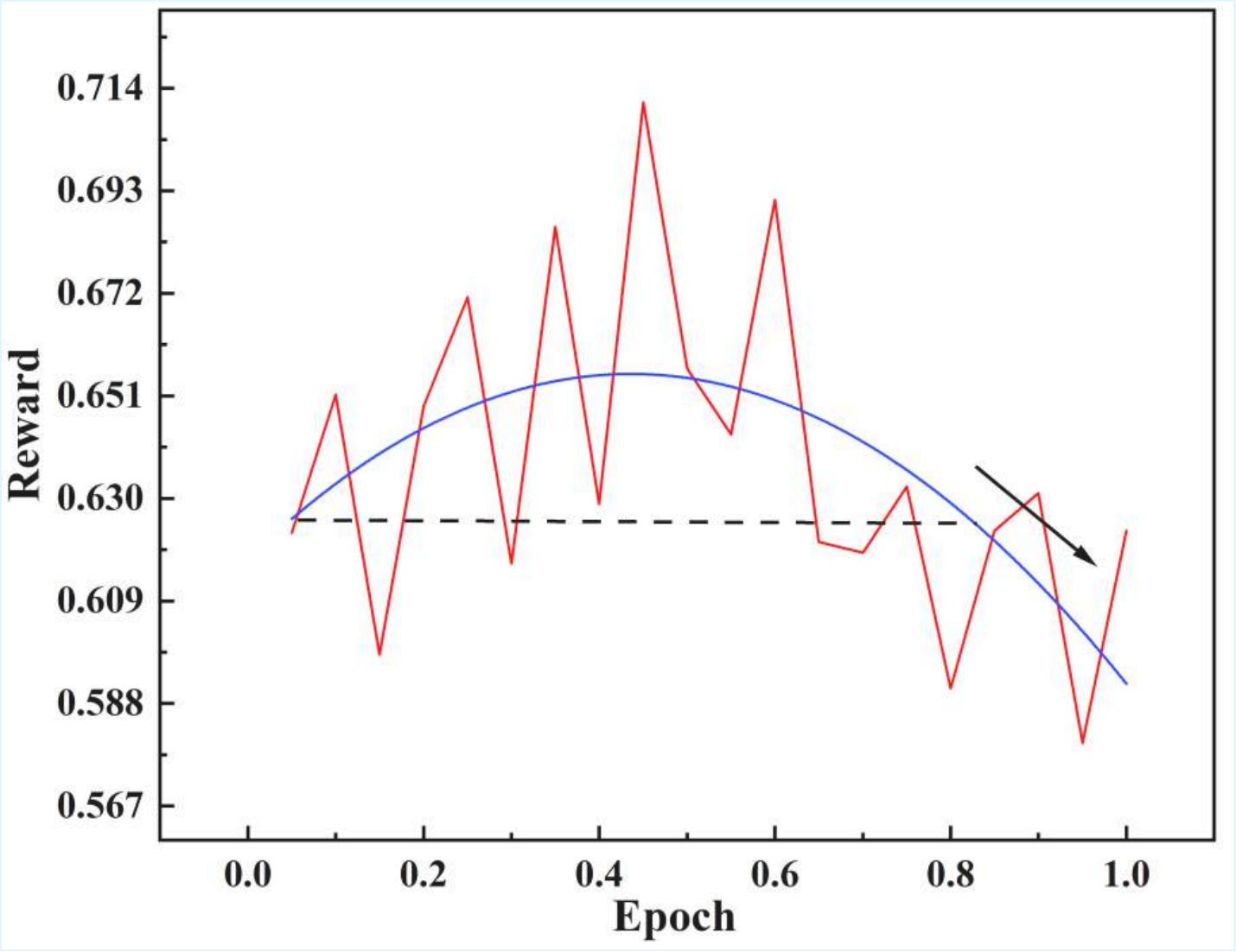}
        \caption{}
        \label{fig:rl_reward_coll}
    \end{subfigure}
    \hfill  
    \begin{subfigure}[t]{0.24\textwidth}
        \centering
        \includegraphics[width=\linewidth]{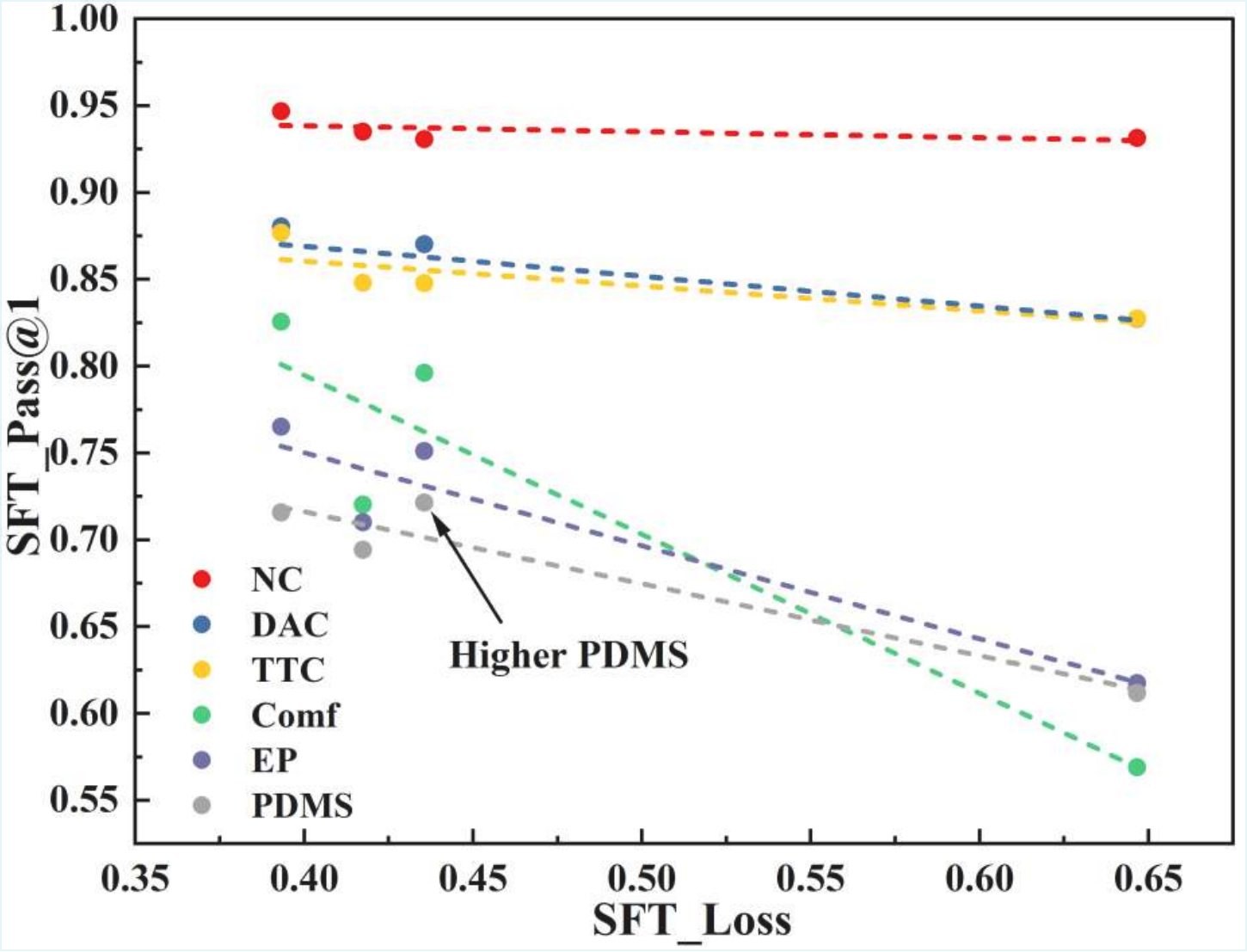}
        \caption{}
        \label{fig:sftloss_sftpass1}
    \end{subfigure}
    \hfill
    \begin{subfigure}[t]{0.24\textwidth}
        \centering
        \includegraphics[width=\linewidth]{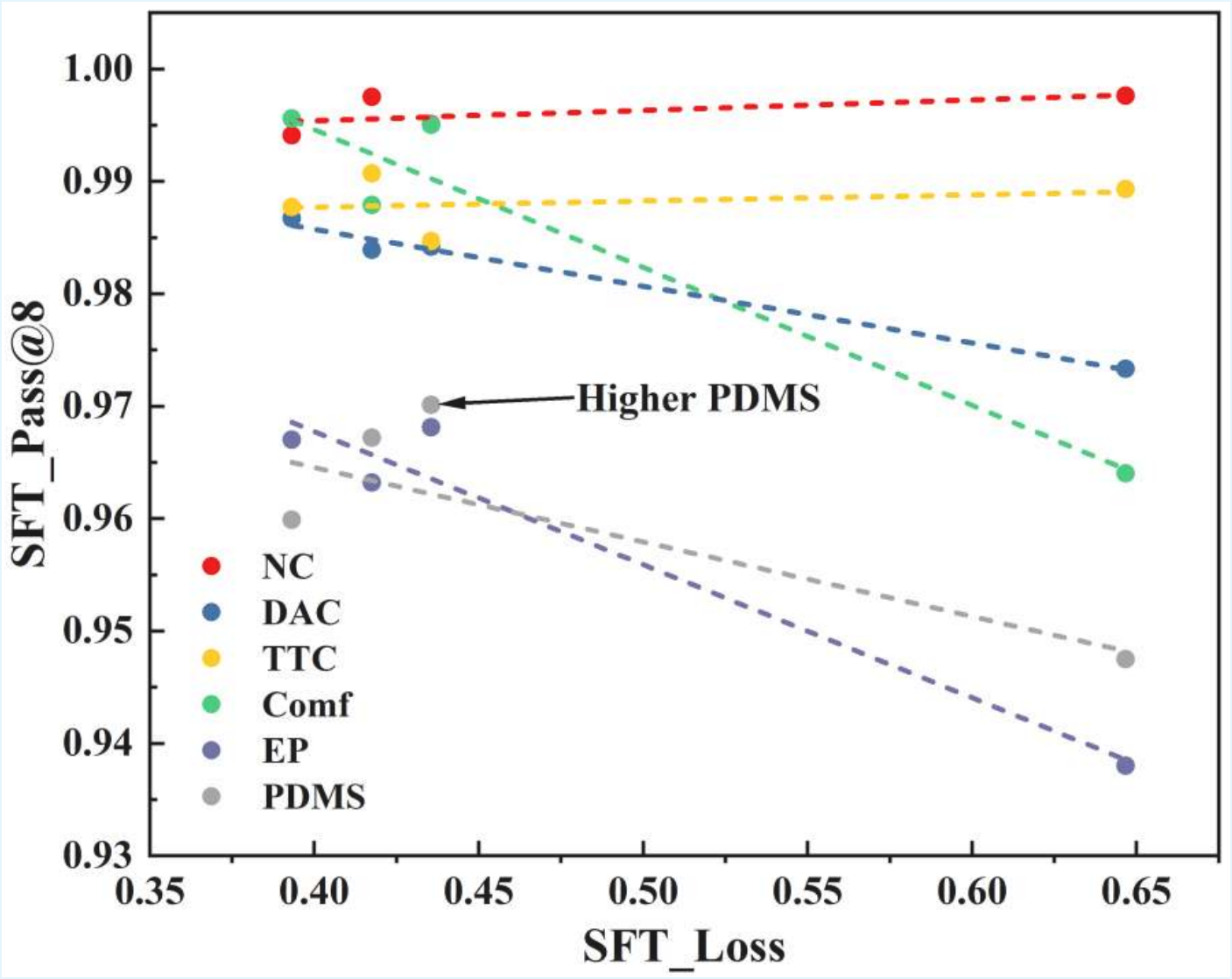}
        \caption{}
        \label{fig:sftloss_sftpass8}
    \end{subfigure}
    \hfill
    \begin{subfigure}[t]{0.24\textwidth}
        \centering
        \includegraphics[width=\linewidth]{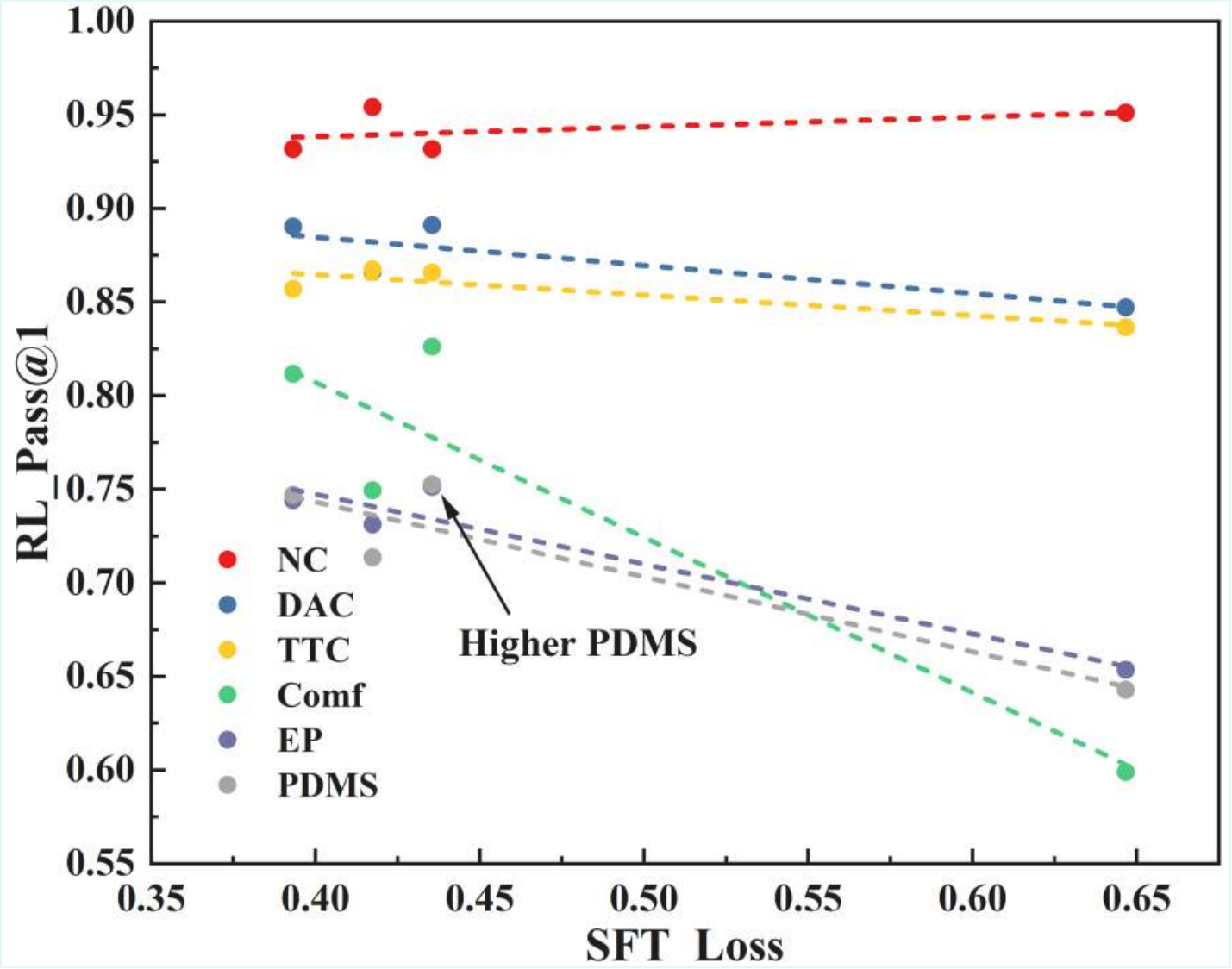}
        \caption{}
        \label{fig:sftloss_RLpass1}
    \end{subfigure}
    
    \caption{Investigate the impact of supervised learning training extent on accuracy, diversity, and RL learning. (a) Selecting an overfitted checkpoint causes the reward signal to fail in providing effective discrimination, which leads to a training collapse. (b) In contrast, a non-overfitted checkpoint achieves a higher performance score during a single inference pass. (c) When aggregating the results of eight inference trials, the non-overfitted checkpoint maintains a higher degree of diversity. (d) Consequently, the checkpoint that demonstrates the best performance in both (b) and (c) yields the highest score after reinforcement learning.}
    \label{fig:ckpt_analyze}
\end{figure*} of downstream trajectory execution, we perform trajectory discretization. Each 2D coordinate $(x, y)$ is mapped to its corresponding 1D codebooks. We define a uniform 1D codebook $\text{Traj} = \{v_1, v_2, \dots, v_K\}$ by discretizing a spatial range $[-M, M]$ with resolution $\Delta g$. A quantizer $Q$ maps a real value to its nearest token, and its inverse $Q^{-1}$ recovers the coordinate. Each 2D waypoint is thus represented by a token pair $(v_x, v_y)$, and the full trajectory of $N$ points becomes a flattened sequence $Q(\tau) = \{(v_{x,n}, v_{y,n})\}_{n=1}^N$. Similarly, other attributes in the planning segment, such as the heading angle $\gamma$, are also mapped to its respective 1D codebooks. To an extent, discrete trajectory tokens hide the precision of the model, which helps mitigate the composite error effects potentially caused by imperfect continuous value predictions. Discretization itself introduces a degree of robustness, preventing noise from redundant tokens from affecting action execution~\cite{seff2023motionlmmultiagentmotionforecasting}.

The \textbf{forward process} gradually corrupts the discretized data $x_0 := x$ into a sequence of increasingly noisy latent variables $x_{1:T} := (x_1, \ldots, x_T)$. 
Formally, it is defined as: $q(x_{1:T} \mid x_0) = \prod_{t=1}^T q(x_t \mid x_{t-1})$. For discrete token modeling with an absorbing state~\cite{Structured-denoising-diffusion-models-in-discrete-state-spaces}(e.g., a special \texttt{[MASK]} token), the forward transition can be formalized as follows: $q(x_t \mid x_{t-1}) = \text{Cat}\!\left(x_t; Q_t^\top x_{t-1}\right)$, where the categorical distribution is parameterized by the transition matrix $Q_t = (1 - \beta_t) I + \beta_t \mathbf{1} m^\top$,
with $m$ denoting the one-hot representation of the absorbing \texttt{[MASK]} token, 
$\mathbf{1}$ a column vector of ones, and $\beta_t \in (0,1)$ the noise schedule. The marginal distribution over $x_t$ conditioned on the original input $x_0$ is then given by: $q(x_t \mid x_0) = \alpha_t x_0 + (1 - \alpha_t) m$, where the coefficient $\alpha_t$ is defined as the product of the survival probabilities up to step $t$: $\alpha_t = \prod_{i=1}^{t} (1 - \beta_i)$.

The \textbf{backward process} is then trained to progressively denoise the latent variables, recovering samples from the data distribution. 
It is defined as: $p_\theta(x) = \sum_{x_{1:T} \sim q} p(x_T) \prod_{t=1}^T p_\theta(x_{t-1} \mid x_t)$, where $p(x_T)$ is typically a simple prior (e.g. uniform distribution), and the model $p_\theta$ parameterizes the reverse denoising transitions.

The loss function used for training is derived from the evidence lower bound (ELBO) and takes the form of a reweighted cross-entropy loss: $\mathcal{L}_{D} = \mathbb{E}_{t} \left[ \mathcal{L}_{t} \right]$, where the per-timestep loss $\mathcal{L}_t$ is defined as:
{
\scriptsize
\begin{equation}
    \mathcal{L}_{t} = \frac{1}{t} \,
    \mathbb{E}_{q(x_t \mid x_0)} \left[
        - \sum_{n=1}^{N} 
        \delta_{x_t^n, m(x_0^n)}^\top 
        \log f_\theta(x_t)^n
    \right].
    \label{eq:discrete-loss-step}
\end{equation}
}

Here, $\delta_{x_t^n, m(x_0^n)}$ is an indicator function that is $1$ if the $n$-th token is masked and $0$ otherwise,  $m(x_0^n)$ denotes the masked version of the original token $x_0^n$, and $f_\theta$ denotes the prediction distribution of model at position $n$ given the corrupted input $x_t$.

\subsection{RL: Diversity Filtering Mechanism}
Supervised Fine-Tuning (SFT) uses expert demonstrations to efficiently shape model behavior, but it may encourage memorization rather than generalization. In contrast, Reinforcement Learning (RL) promotes active exploration and improves policies through reward feedback, potentially discovering solutions beyond expert trajectories. However, if SFT overfits the training data and converges to a local optimum, the model may lose exploration capability during RL, resulting in policy degradation~\cite{zhang2026onpolicyrlmeetsoffpolicy,shao2024deepseekmathpushinglimitsmathematical}. In driving tasks, we observe policy distribution collapse during RL: as shown in Fig.~\ref{fig:rl_reward_coll}, highly homogeneous rollout samples reduce the discriminative power of reward signals, causing policy exploration to stagnate.

So, the filtering mechanism first establishes a candidate set of SFT policies, denoted as $\mathcal{P}_{cand} = \{\pi_1, \pi_2, \dots, \pi_n\}$. To evaluate these candidates, we employ the $\text{pass@1}$ metric to measure accuracy, defined as $\text{Acc}_i = \text{pass@1}(\pi_i)$, and the $\text{pass@}k$ metric (where $n$ represents the total sample budget) to quantify the exploration diversity, expressed as $\text{Div}_i = \text{pass@}n(\pi_i)$. The optimal policy $\pi^*$ is then identified by satisfying the following criteria for all $j \in \{1, \dots, n\}$: $$
\left|\text{Acc}_{\pi^*}-\text{Acc}_j\right| \le \epsilon
\quad \text{and} \quad
\text{Div}_{\pi^*} \ge \text{Div}_j
$$By applying this joint constraint, we select the policy that maximizes both execution success and state-space coverage, thereby providing a superior starting point for subsequent reinforcement learning. 

As shown in Fig.~\ref{fig:ckpt_analyze}(b) to (d), the checkpoint performs well under relatively high sftloss conditions, achieving good accuracy and diversity scores for pass@1 and pass@8 in the dense evaluation before RL. After RL, the same checkpoint, which has already demonstrated good performance, achieved the highest PDMS. This proves that ensuring diversity and accuracy in SFT is an effective way to improve learning efficiency through RL, and also demonstrates that a filtering mechanism is necessary during the RL process.

We also implement a heuristic filtering strategy for the reinforcement learning dataset. We define three distinct difficulty levels: $\mathcal{D} = \{\text{easy}, \text{mid}, \text{hard}\}$. For each query in the total data distribution $P(Q)$, we perform $n$ stochastic rollouts. Each sample is then assigned to a specific difficulty interval based on the mode of its rollout outcomes (i.e., the most frequently occurring difficulty level among the $n$ trials). Formally, we define the refined training distribution $P(Q)^*$ by selecting only the samples categorized as mid or hard. By excluding easy samples, which provide limited gradients for policy improvement, we focus the reinforcement learning process on challenging scenarios to maximize the efficiency of strategy optimization. Here is the updated GRPO objective function of the training, where the policy sampling and optimization are now initialized from the optimal policy $\pi^*$ (the result of our diversity filtering mechanism):
{\scriptsize
\begin{equation}
\begin{aligned}
J(\theta) ={}&
\mathbb{E}_{q \sim P(Q)^*,\, \{o_i\}_{i=1}^G \sim \pi^*_{\mathrm{old}}}
\Bigg[
\frac{1}{G}\sum_{i=1}^G
\frac{1}{|o_i|}
\sum_{k=1}^{|o_i|}
\min\Bigg(
\frac{\pi_\theta(o_i^k \mid q,o_{i,t<T}^k)}
{\pi_{\mathrm{old}}^*(o_i^k \mid q,o_{i,t<T}^k)}A_i,
\\
&\qquad
\operatorname{clip}\left(
\frac{\pi_\theta(o_i^k \mid q,o_{i,t<T}^k)}
{\pi_{\mathrm{old}}^*(o_i^k \mid q,o_{i,t<T}^k)},
1-\varepsilon,
1+\varepsilon
\right)A_i
\Bigg)
-\beta D_{\mathrm{KL}}(\pi_\theta \Vert \pi_{\mathrm{ref}})
\Bigg].
\end{aligned}
\end{equation}
}

\textbf{Theoretical Analysis:} An over-converged (overfitted) SFT checkpoint produces nearly identical rollouts, which weakens the learning signal of GRPO. Specifically, when rollout diversity collapses, normalized advantage $\hat{A}_i=\frac{r_i-\bar r}{\sigma_r+\epsilon}$ and policy gradient $\nabla_\theta J \propto \sum_{i=1}^{G} \hat{A}_i \nabla_\theta \log \pi_\theta(o_i\mid q)$ both diminish. Under such conditions, $r_i \approx r_j$ for all $i,j$, leading to $\sigma_r \approx 0$, $\hat{A}_i \approx 0$, and consequently $\nabla_\theta J \approx 0$. In contrast, a non-overfitted checkpoint preserves recoverable solution modes, reflected by a larger $\operatorname{pass@K}$. This produces more discriminative reward signals and more effective RL updates, ultimately improving the final policy.

\subsection{Inference: Dynamic Autonomous Driving System Decoding Strategy}

Traditional decoding relies on token-level confidence for sampling, prioritizing high-confidence tokens. Under this mechanism, the system is prone to frequent switching between disparate driving modes, introducing significant safety uncertainty into the Autonomous Driving System (ADS). To address this, we propose the Dynamic Autonomous Driving decoding strategy.

Given a predicted token sequence of length $N$, let $\boldsymbol{\ell}_i$ denote the output logits at the $i$-th position. The confidence of the $i$-th token is defined as
{
\scriptsize
\begin{equation}
C_i = \max_{v \in \mathcal{V}}
\operatorname{Softmax}(\boldsymbol{\ell}_i)_v,
\end{equation}
}

where $\mathcal{V}$ denotes the token vocabulary. We then compute the cumulative confidence scores for the proximal and distal segments:
{
\scriptsize
\begin{equation}
\Phi_{\mathrm{prox}}
=
\sum_{i=1}^{\lfloor N/2 \rfloor} C_i,
\qquad
\Phi_{\mathrm{dist}}
=
\sum_{i=\lfloor N/2 \rfloor + 1}^{N} C_i.
\end{equation}
}

The decoding mode $\mathcal{D}$ is then determined by the following decision logic:
{
\scriptsize
\begin{equation}
\mathcal{D} =
\begin{cases}
\text{Forward (Near-to-Far)}, &
\text{if } \Phi_{\mathrm{prox}} \geq \Phi_{\mathrm{dist}}, \\[2mm]
\text{Reverse (Far-to-Near)}, &
\text{if } \Phi_{\mathrm{prox}} < \Phi_{\mathrm{dist}}.
\end{cases}
\end{equation}
}

Accordingly, the decoding order is formally defined as
{
\scriptsize
\begin{equation}
\pi_{\mathcal{D}} =
\begin{cases}
(1,2,\ldots,N), &
\mathcal{D}=\text{Near-to-Far}, \\[1mm]
(N,N-1,\ldots,1), &
\mathcal{D}=\text{Far-to-Near}.
\end{cases}
\end{equation}
}

At inference step $k$, let $\mathcal{S}_k$ denote the set of currently masked positions. Given two consecutive timesteps $t_k$ and $t_{k+1}$, the number of tokens to be decoded is
{
\scriptsize
\begin{equation}
n_k =
\left\lceil
|\mathcal{S}_k|
\left(
1-\frac{t_{k+1}}{t_k}
\right)
\right\rceil.
\end{equation}
}

The first $n_k$ masked positions following the order $\pi_{\mathcal{D}}$ are selected and replaced by their predicted tokens. The selected decoding mode remains fixed throughout the inference process, thereby avoiding frequent switching between different generation directions.

The distribution area of the high-confidence points is calculated to determine the decoding order. In low-risk scenarios, such as smooth cruising with no leading vehicles, the decision-making weight shifts toward comfort. In this case, the Near-to-Far forward inference mode is adopted to ensure trajectory smoothness and passenger experience. Conversely, in high-risk scenarios, such as car-following in complex traffic environments, driving decisions follow a safety-first cognitive logic based on the anchor prediction principle~\cite{zhang2020mapadaptivegoalbasedtrajectoryprediction}. The system first locks the long-term furthest safety point and then employs a Far-to-Near reverse inference mode to determine internal trajectory tokens, thereby ensuring the global safety of the path.

\section{Experiments}
\textbf{Implementation:} All trainings are conducted on four NVIDIA H800 GPUs. Training is performed in bf16 precision. We employ DreamVL~\cite{ye2026dreamvldreamvlaopen} as the foundation for our\begin{table}[t]
  \centering
  \small
  \setlength{\tabcolsep}{2pt}
  \renewcommand{\arraystretch}{1.05}
  \begin{adjustbox}{max width=\columnwidth}
  \begin{tabular}{@{}lcccccccc@{}}
    \toprule
    \textbf{Method}
    & \textbf{C}
    & \textbf{L}
    & \textbf{NC} $\uparrow$
    & \textbf{DAC} $\uparrow$
    & \textbf{TTC} $\uparrow$
    & \textbf{Comf.} $\uparrow$
    & \textbf{EP} $\uparrow$
    & \textbf{PDMS} $\uparrow$ \\
    \midrule

    \multicolumn{9}{@{}l}{\textit{End-to-end methods}} \\
    TransFuser~\cite{TransFuser}
    & \ding{51} & \ding{51} & 97.7 & 92.8 & 92.8 & 100 & 79.2 & 84.0 \\
    UniAD~\cite{hu2023_uniad}
    & \ding{51} & \ding{55} & 97.8 & 91.9 & 92.9 & 100 & 78.8 & 83.4 \\
    VADv2~\cite{jiang2026vadv2}
    & \ding{51} & \ding{55} & 97.2 & 89.1 & 91.6 & 100 & 76.0 & 80.9 \\
    PARA-Drive~\cite{PARA-Drive}
    & \ding{51} & \ding{55} & 97.9 & 92.4 & 93.0 & 99.8 & 79.3 & 84.0 \\
    Hydra-MDP-V~\cite{li2024hydramdp}
    & \ding{51} & \ding{51} & 97.9 & 91.7 & 92.9 & 100 & 77.6 & 83.0 \\
    DRAMA~\cite{yuan2024drama}
    & \ding{51} & \ding{51} & 98.0 & 93.1 & 94.8 & 100 & 80.1 & 85.5 \\
    DiffusionDrive-Cam~\cite{diffusiondrive}
    & \ding{51} & \ding{55} & 97.8 & 92.2 & 92.6 & 99.9 & 78.9 & 83.6 \\

    \midrule
    \multicolumn{9}{@{}l}{\textit{Autoregressive vision-language models}} \\
    DrivingGPT~\cite{chen2024drivinggptunifyingdrivingworld}
    & \ding{51} & \ding{55} & 98.9 & 90.7 & 94.9 & 95.6 & 79.7 & 82.4 \\
    FSDrive~\cite{zeng2025futuresightdrive}
    & \ding{51} & \ding{55} & 98.2 & 93.8 & 93.3 & 99.9 & 80.1 & 85.1 \\
    QwenVL2.5$\dagger$
    & \ding{51} & \ding{55} & 97.8 & 92.1 & 92.8 & 100 & 78.3 & 83.3 \\
    InternVL3$\dagger$
    & \ding{51} & \ding{55} & 97.0 & 92.4 & 91.8 & 100 & 78.9 & 83.3 \\

    \midrule
    \multicolumn{9}{@{}l}{\textit{Discrete Diffusion vision-language models}} \\
    \rowcolor{gray!10}
    \textbf{MD-Drive (Ours)}
    & \ding{51} & \ding{55}
    & 97.1 & 94.0 & 92.2 & 99.2
    & \textbf{90.3} & \textbf{88.9} \\
    \bottomrule
  \end{tabular}
  \end{adjustbox}
  \caption{Comparison with standard VLM and AR generation methods on NAVSIM-v1. C: camera(s); L: LiDAR. $\dagger$ is from ~\cite{li2025recogdrive}}
  \label{tab:navsim_comparison}
\end{table}MD-Drive, which consists of three primary components: a vision encoder, a connector, and a language backbone. During the trajectory discretization in the SFT stage, we set the $x$-coordinate range to [-5, 70] with a resolution $\Delta g$ of 0.25, the $y$-coordinate range to [-20, 25] with a resolution $\Delta g$ of 0.25, and the steering angle $\gamma$ range to [-2, 2] with a resolution $\Delta g$ of 0.04. In the RL stage, the number of responses per rollout group is 8, and the LoRA rank is set to 8. During the inference stage, the number of iterations for response generation is 32 steps, with the temperature set to 1 and top\_p set to 1.
\begin{table}[t]
\centering
\small
\setlength{\tabcolsep}{2.2pt}
\renewcommand{\arraystretch}{1.08}

\begin{adjustbox}{max width=\columnwidth}
\begin{tabular}{@{}lcccccc@{}}
    \toprule
    \textbf{Method}
    & \textbf{Arch.}
    & \textbf{Gen.}
    & \shortstack{\textbf{SFT}\\\textbf{Data}}
    & \shortstack{\textbf{RL}\\\textbf{Data}}
    & \textbf{PDMS}
    & \shortstack{\textbf{DPP} $\downarrow$\\\textbf{(K/pt.)}} \\
    \midrule

    ReCogDrive~\cite{li2025recogdrive}
    & VLM+Action Expert
    & FM
    & 3100K
    & 85K
    & 90.8
    & 35.08 \\

    WAM-Diff~\cite{xu2025wamDIFF}
    & VLM-MoE
    & DD
    & 1468K
    & 103K
    & 91.0
    & 17.26 \\

    WAM-Flow~\cite{xu2025wamFLOW}
    & VLM
    & DFM
    & 668K
    & 103K
    & 90.3
    & 8.54 \\

    DriveVLA-W0~\cite{li2025drivevlaW0}
    & WM
    & FM
    & 70000K
    & 85K
    & 91.0
    & 770.16 \\

    Epona~\cite{zhang2025epona}
    & WM
    & AR+RF
    & N/R$^\dagger$
    & --
    & 86.2
    & -- \\

    ViLaD~\cite{cui2025viladlargevisionlanguage}
    & VLM
    & DD
    & --
    & --
    & --
    & -- \\

    MVLAD-AD~\cite{zhang2026efficientexplainableendtoendautonomous}
    & VLM
    & DD
    & --
    & --
    & --
    & -- \\

    \midrule
    \rowcolor{gray!10}
    \textbf{MD-Drive}
    & \textbf{VLM}
    & \textbf{DD}
    & \textbf{153K}
    & \textbf{58K}
    & \textbf{88.9}
    & \textbf{2.37} \\

    \bottomrule
\end{tabular}
\end{adjustbox}

\parbox{\columnwidth}{\scriptsize
DPP is calculated as
$(V_{\mathrm{SFT}}+V_{\mathrm{RL}})/\mathrm{PDMS}$,
with all reported data volumes converted to thousands of samples.
AR: Autoregressive; DFM: Discrete Flow Matching;
FM: flow matching; DD: Discrete Diffusion;
RF: Rectified Flow; WM: World Model.
$^\dagger$Epona reports training on public nuPlan videos and
700 nuScenes scenes, but does not provide a directly comparable
sample count. N/R denotes not reported.
}
\caption{Comparison of methods employing non-standard VLM architectures, non-autoregressive generation strategies, or different training data scales on NAVSIM-v1. DPP denotes the amount of training data per performance point, with lower values indicating higher data efficiency.}
\label{tab:navsim_comparison2}
\end{table}

\subsection{Comparision With Existed Works on NAVSIM}
Under a comparable standard VLM architecture, without introducing an additional action expert such as a diffusion planner or adopting a mixture-of-experts transformer, MD-Drive (Discrete Diffusion Generation) demonstrates a clear advantage over autoregressive generation. Specifically, as shown in~\ref{tab:navsim_comparison}, MD-Drive achieves a PDMS of 88.9, outperforming DrivingGPT and FSDrive by 6.5 and 3.8 points, respectively. These results indicate that discrete diffusion provides a more effective trajectory generation paradigm than conventional autoregressive decoding under a similarly streamlined architectural setting. Meanwhile, MD-Drive achieves a throughput of 243.9 tokens/s and an inference frequency of 13.35 Hz, while incurring only a 2.0-point performance drop.

As shown in Table~\ref{tab:navsim_comparison2}, MD-Drive achieves competitive performance with substantially less training data than existing methods employing action experts, MoE architectures, or world models. Although its PDMS is slightly lower than that of WAM-Flow, WAM-Diff, and ReCogDrive, MD-Drive attains the lowest DPP of 2.37, compared with 8.54, 17.26, and 35.08, respectively. In particular, it is approximately 3.6$\times$ more data-efficient than WAM-Flow, demonstrating that discrete diffusion can provide a favorable balance between driving performance and training cost.

\subsection{Comparision With Existed Works on nuScenes}
As shown in Table~\ref{tab:nuscenes_comparison}, discrete-diffusion VLMs generally outperform autoregressive VLMs in both trajectory accuracy and failure rate. MD-Drive achieves the best average L2 error of 0.78, improving over the strongest autoregressive baseline by 1.68 and surpassing MVLAD-AD, the best competing discrete-diffusion method, by 0.50. Moreover, MD-Drive reduces the failure rate to 0.00\%, demonstrating that trajectory discretization enables more accurate and reliable trajectory generation under a standard VLM architecture.

\begin{table}[t]
  \centering

  \small
  \setlength{\tabcolsep}{2.2pt}
  \renewcommand{\arraystretch}{1.05}

  \begin{adjustbox}{max width=\columnwidth}
    \begin{tabular}{@{}lllccccc@{}}
      \toprule
      \textbf{Method}
      & \textbf{Mod.}
      & \textbf{Backbone}
      & \textbf{L2-1s} $\downarrow$
      & \textbf{L2-2s} $\downarrow$
      & \textbf{L2-3s} $\downarrow$
      & \textbf{Avg.} $\downarrow$
      & \textbf{FR} $\downarrow$ \\
      \midrule

      \multicolumn{8}{@{}l}{\textit{Autoregressive VLMs}} \\
      LLaVA-1.6-7B (Z)$^\dagger$
      & 1$\times$C & LLaVA
      & 1.66 & 3.54 & 4.54 & 3.24 & 4.06 \\

      Llama-3.2-11B (Z)$^\dagger$
      & 1$\times$C & Llama
      & 1.50 & 3.44 & 4.04 & 3.00 & 23.9 \\

      Qwen2-VL-7B (Z)$^\dagger$
      & 1$\times$C & Qwen
      & 1.22 & 2.94 & 3.21 & 2.46 & 24.0 \\

      LLaVA-1.6 (O-E)$^\dagger$
      & 1$\times$C & LLaVA
      & 1.49 & 3.38 & 4.09 & 2.98 & 6.12 \\

      Llama-3.2 (O-E)$^\dagger$
      & 1$\times$C & Llama
      & 1.54 & 3.31 & 3.91 & 2.92 & 22.0 \\

      Qwen2-VL (O-E)$^\dagger$
      & 1$\times$C & Qwen
      & 1.45 & 3.21 & 3.76 & 2.81 & 16.1 \\

      \midrule
      \multicolumn{8}{@{}l}{\textit{Discrete-diffusion VLMs}} \\
      ViLaD-Zero$^\ddagger$
      & 1$\times$C & LLaDA-V
      & 1.09 & 2.51 & 3.74 & 2.45 & 0.03 \\

      ViLaD-CoT$^\ddagger$
      & 1$\times$C & LLaDA-V
      & 1.24 & 2.74 & 4.13 & 2.71 & 0.17 \\

      ViLaD-SFT$^\ddagger$
      & 1$\times$C & LLaDA-V
      & 0.79 & 1.92 & 2.83 & 1.85 & 0.00 \\

      ViLaD-Opt$^\ddagger$
      & 1$\times$C & LLaDA-V
      & 0.81 & 1.93 & 2.69 & 1.81 & 0.00 \\

      MVLAD-AD~\cite{zhang2026efficientexplainableendtoendautonomous}
      & 1$\times$C & LLaDA-V
      & 0.70 & 1.31 & 2.34 & 1.28 & 0.00 \\

      DreamVL-Zero
      & 1$\times$C & DreamVL
      & 4.62 & 7.75 & 10.9 & 7.75 & 23.0 \\

      \rowcolor{gray!10}
      \textbf{MD-Drive (Ours)}
      & 1$\times$C & DreamVL
      & \textbf{0.42}
      & \textbf{0.71}
      & \textbf{1.21}
      & \textbf{0.78}
      & \textbf{0.00} \\

      \bottomrule
    \end{tabular}
  \end{adjustbox}

  \caption{Comparison between autoregressive and discrete-diffusion generation paradigms using standard VLM architectures on nuScenes. Lower values indicate better performance. $^\dagger$ Results are from~\cite{openemma}, and $^\ddagger$ results are from~\cite{cui2025viladlargevisionlanguage}. Z: zero-shot; O-E: OpenEMMA; C: camera.}
  \label{tab:nuscenes_comparison}
\end{table}

\begin{table}[t]
  \centering

  \small
  \setlength{\tabcolsep}{1.8pt}
  \renewcommand{\arraystretch}{1.08}

  \begin{adjustbox}{max width=\columnwidth}
    \begin{tabular}{@{}lccccccccc@{}}
      \toprule
      \textbf{Method}
      & \textbf{C}
      & \textbf{L}
      & \textbf{SL}
      & \textbf{FA}
      & \textbf{SD}
      & \textbf{SA}
      & \textbf{CT}
      & \textbf{BR}
      & \textbf{W-F1} \\
      \midrule

      TransFuser\ding{61}
      & \ding{51} & \ding{51}
      & -- & -- & -- & -- & -- & 0.10 & 0.10 \\

      TCP\ding{61}
      & \ding{51} & \ding{55}
      & -- & -- & -- & -- & -- & 0.21 & 0.21 \\

      InterFuser\ding{61}
      & \ding{51} & \ding{51}
      & -- & -- & -- & -- & -- & 0.35 & 0.35 \\

      DriveCoT-Agent\ding{61}
      & \ding{51} & \ding{55}
      & \textbf{0.87}
      & \textbf{0.81}
      & \textbf{0.75}
      & 0.72
      & \textbf{0.83}
      & \textbf{0.84}
      & \textbf{0.82} \\

      \midrule

      Qwen2.5-VL-7B (SFT)
      & \ding{51} & \ding{55}
      & 0.80
      & 0.73
      & 0.48
      & 0.68
      & 0.39
      & 0.72
      & 0.73 \\

      \rowcolor{gray!10}
      \textbf{MD-Drive (Ours)}
      & \ding{51} & \ding{55}
      & \textbf{0.87}
      & 0.80
      & 0.55
      & \textbf{0.79}
      & 0.47
      & \textbf{0.84}
      & \textbf{0.81} \\

      \bottomrule
    \end{tabular}
  \end{adjustbox}

  \caption{Comparison of driving behavior prediction performance in terms of F1-score. Results marked with \ding{61} are from~\cite{wang2024drivecot}. C: camera(s); L: LiDAR; SL: Speed Limit; FA: Follow Ahead; SD: Slow Down; SA: Slow Approach; CT: Cautious Turn; BR: Brake; W-F1: weighted F1-score.}
  \label{tab:drivecot_comparison}
\end{table}

\subsection{Comparision With Existed Works on DriveCoT}
As shown in Table~\ref{tab:drivecot_comparison}, MD-Drive achieves a weighted F1-score of 0.81, outperforming the autoregressively SFT-trained Qwen2.5-VL-7B by 0.08. It performs comparably to the end-to-end DriveCoT-Agent, with only a 0.01 gap in weighted F1-score. This slight difference is partly attributed to the ambiguity of driving-behavior annotations, as multiple actions may be valid in the same scenario; for example, a vehicle may either slow down or brake in response to a potential hazard. We provide further discussion of this issue in the Appendix.

\subsection{Ablation Studies}
\begin{figure*}[htbp]
    \centering
    \begin{subfigure}[t]{0.40\textwidth}
        \centering
        \includegraphics[width=\linewidth]{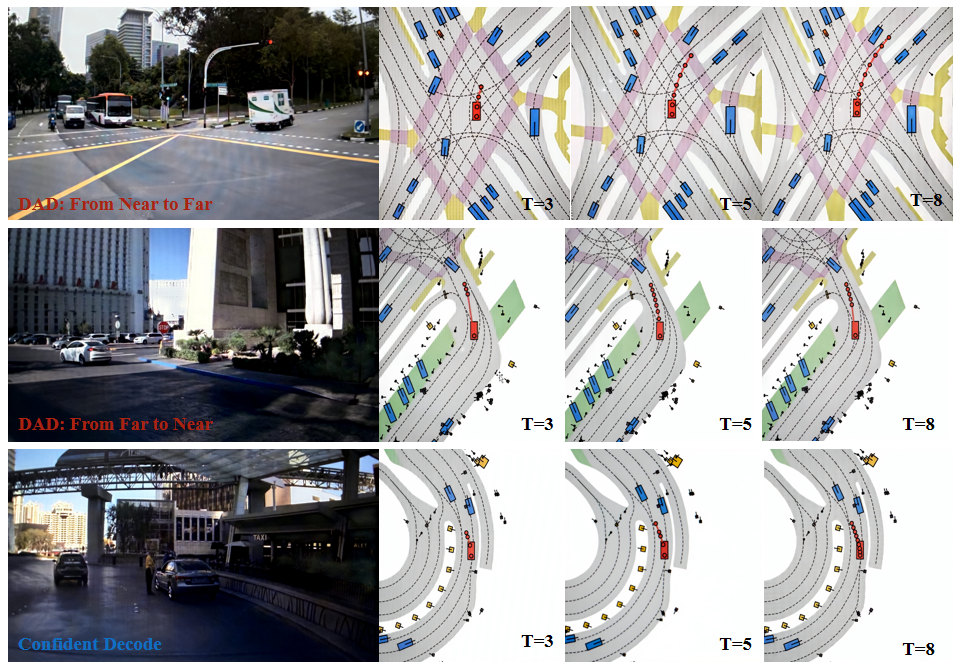}
        \caption{}
        \label{fig:decode_example}
    \end{subfigure}
    \hspace{0.015\textwidth}
    \begin{subfigure}[t]{0.35\textwidth}
        \centering
        \includegraphics[width=\linewidth]{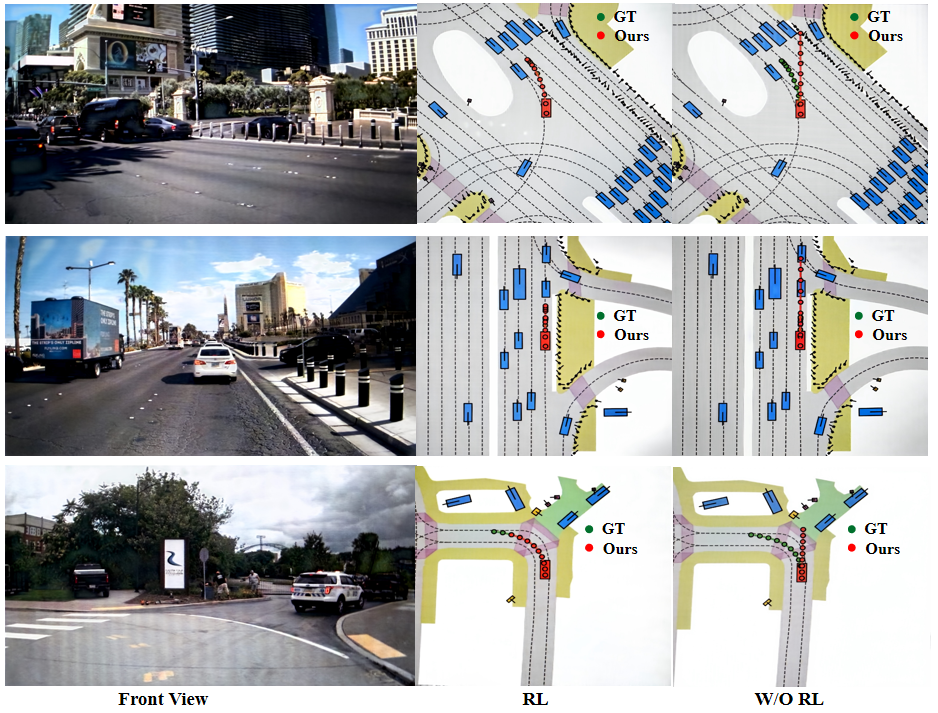}
        \caption{}
        \label{fig:rl_example}
    \end{subfigure}

    \caption{(a) Illustration of different decoding strategies.
    (b) Comparison of different scenarios before and after GRPO training.}
    \label{fig:two-subfigures}
\end{figure*}

As shown in Table~\ref{tab:ablation_navsim}, each stage of the proposed optimization method consistently improves model performance, confirming the effectiveness of trajectory discretization, DAD decoding, and GRPO. The non-overfitted and overfitted settings correspond to Checkpoint-5318 and Checkpoint-10636, respectively. As further shown in Table~\ref{tab:multi_seed}, Checkpoint-5318 achieves a higher pass@8 score than Checkpoint-10636 (96.72\% vs.\ 95.99\%) and yields a substantially higher PDMS after GRPO (88.9 vs.\ 85.8). This result suggests that checkpoints with stronger multi-seed pass@8 stability can be optimized more effectively during GRPO, even when their average PDMS score is lower. Fig.~\ref{fig:two-subfigures}(b) visualizes the trajectory prediction results across different scenarios before and after GRPO.\begin{table}[t]
  \centering
  \small
  \setlength{\tabcolsep}{1.55pt}
  \renewcommand{\arraystretch}{1.05}

  \begin{adjustbox}{max width=\columnwidth}
    \begin{tabular}{@{}cccccccccc@{}}
      \toprule
      \textbf{ID}
      & \textbf{Disc.}
      & \textbf{DAD}
      & \textbf{GRPO}
      & \textbf{NC} $\uparrow$
      & \textbf{DAC} $\uparrow$
      & \textbf{TTC} $\uparrow$
      & \textbf{Comf.} $\uparrow$
      & \textbf{EP} $\uparrow$
      & \textbf{PDMS} $\uparrow$ \\
      \midrule

      \multicolumn{10}{@{}l}{\textit{Non-overfitted checkpoint}} \\
      1
      & \ding{55} & \ding{55} & \ding{55}
      & 91.8 & 80.9 & 84.0 & 95.9 & 67.6 & 68.5 \\

      2
      & \ding{51} & \ding{55} & \ding{55}
      & 94.7 & 87.9 & 87.3 & 81.7 & 80.0 & 76.9 \\

      3
      & \ding{51} & \ding{51} & \ding{55}
      & 96.4 & 87.4 & 90.1 & 93.1 & 82.0 & 80.3 \\

      \rowcolor{gray!10}
      4
      & \ding{51} & \ding{51} & \ding{51}
      & \textbf{97.1}
      & \textbf{94.0}
      & \textbf{92.2}
      & \textbf{99.2}
      & \textbf{90.3}
      & \textbf{88.9} \\

      \midrule
      \multicolumn{10}{@{}l}{\textit{Overfitted checkpoint}} \\
      1
      & \ding{55} & \ding{55} & \ding{55}
      & 91.8 & 80.9 & 84.0 & 95.9 & 67.6 & 68.5 \\

      2
      & \ding{51} & \ding{55} & \ding{55}
      & 97.1 & 87.7 & 91.3 & 95.0 & 82.3 & 81.4 \\

      3
      & \ding{51} & \ding{51} & \ding{55}
      & \textbf{97.3}
      & 90.6
      & \textbf{91.7}
      & \textbf{96.8}
      & 84.6
      & 84.3 \\

      \rowcolor{gray!10}
      4
      & \ding{51} & \ding{51} & \ding{51}
      & 96.0
      & \textbf{92.9}
      & 89.4
      & 96.2
      & \textbf{88.3}
      & \textbf{85.8} \\

      \bottomrule
    \end{tabular}
  \end{adjustbox}

  \caption{Ablation study of the multi-stage optimization pipeline using non-overfitted and overfitted checkpoints.}
  \label{tab:ablation_navsim}
\end{table}

\begin{table}[t]
    \centering
    \small
    \setlength{\tabcolsep}{4pt}
    \renewcommand{\arraystretch}{1.08}

    \begin{adjustbox}{max width=\columnwidth}
        \begin{tabular}{@{}lcccc@{}}
            \toprule
            \textbf{Checkpoint}
            & \shortstack{\textbf{PDMS Mean $\pm$ Std.} $\uparrow$\\
                          \textbf{(8 steps Infer)}} 
            & \textbf{pass@1} $\uparrow$
            & \textbf{pass@8} $\uparrow$
            & \shortstack{\textbf{PDMS after GRPO} $\uparrow$\\
                          \textbf{(32 steps)}} \\
            \midrule

            Checkpoint-2659
            & 65.71 $\pm$ 0.26
            & 61.18\%
            & 94.75\%
            & 85.9 $\pm$ 0.13 \\

            \rowcolor{gray!10}
            Checkpoint-5318
            & 71.64 $\pm$ 0.30
            & 69.42\%
            & \textbf{96.72\%}
            & \textbf{88.9 $\pm$ 0.08} \\

            Checkpoint-7977
            & 74.40 $\pm$ 0.30
            & 71.13\%
            & 96.42\%
            & 86.4 $\pm$ 0.12 \\

            Checkpoint-10636
            & \textbf{77.20 $\pm$ 0.23}
            & \textbf{71.56\%}
            & 95.99\%
            & 85.8 $\pm$ 0.09 \\

            \bottomrule
        \end{tabular}
    \end{adjustbox}

    \caption{Multi-seed stability analysis for checkpoint selection. Each checkpoint is evaluated using eight random seeds.}
    \label{tab:multi_seed}
\end{table}

\begin{table}[t]
    \centering
    \small
    \setlength{\tabcolsep}{1.8pt}
    \renewcommand{\arraystretch}{1.06}

    \begin{adjustbox}{max width=\columnwidth}
        \begin{tabular}{@{}lcccccccc@{}}
            \toprule
            \textbf{Decoding}
            & \textbf{Heur.}
            & \shortstack{\textbf{Var.}\\\textbf{Order}}
            & \textbf{NC} $\uparrow$
            & \textbf{DAC} $\uparrow$
            & \textbf{TTC} $\uparrow$
            & \textbf{Comf.} $\uparrow$
            & \textbf{EP} $\uparrow$
            & \shortstack{\textbf{PDMS} $\uparrow$\\\textbf{(32 steps)}} \\
            \midrule

            Confidence-only
            & \ding{51}
            & \ding{53}
            & 97.3
            & 90.6
            & 91.7
            & 96.8
            & 84.6
            & 84.3 \\

            Causal$^\ddagger$
            & \ding{51}
            & \ding{53}
            & 97.1
            & 87.7
            & 91.3
            & 95.0
            & 82.3
            & 81.4 \\

            R-Causal$^\ddagger$
            & \ding{51}
            & \ding{53}
            & \textbf{97.4}
            & 92.2
            & \textbf{92.3}
            & 96.2
            & 87.6
            & 86.5 \\

            \rowcolor{gray!10}
            \textbf{DAD (Ours)}
            & \ding{51}
            & \ding{51}
            & 97.1
            & \textbf{94.0}
            & 92.2
            & \textbf{99.2}
            & \textbf{90.3}
            & \textbf{88.9} \\

            \bottomrule
        \end{tabular}
    \end{adjustbox}

    \caption{Ablation study of different decoding strategies on NAVSIM. Heur. denotes heuristic decoding, and Var. Order denotes a dynamically adjustable decoding order during continuous inference. $^\ddagger$ Results are from~\cite{xu2025wamDIFF}.}
    \label{tab:dad_ablation}
\end{table}

\begin{table}[t]
    \centering
    \small
    \setlength{\tabcolsep}{1.5pt}
    \renewcommand{\arraystretch}{1.06}

    \begin{adjustbox}{max width=\columnwidth}
        \begin{tabular}{@{}lccccc@{}}
            \toprule
            $\boldsymbol{\Delta g}$
            & \shortstack{\textbf{Vocab.}\\\textbf{Size}}
            & \shortstack{\textbf{Quant.}\\\textbf{Error (m)}}
            & \textbf{SFT}
            & \textbf{RL}
            & \textbf{PDMS} $\uparrow$ \\
            \midrule

            No Disc.
            & --
            & --
            & \ding{51}
            & \ding{53}
            & 68.5 \\

            0.5
            & +340
            & 0.25
            & \ding{51}
            & \ding{53}

            & 78.2 \\

            \rowcolor{gray!10}
            0.25
            & +580
            & 0.125
            & \ding{51}
            & \ding{53}
            & \textbf{81.4} \\

            0.1
            & +1300
            & 0.05
            & \ding{51}
            & \ding{53}
            & 80.8 \\

            \bottomrule
        \end{tabular}
    \end{adjustbox}

    \caption{Sensitivity analysis of trajectory discretization on NAVSIM.}
    \label{tab:discretization_sensitivity}
\end{table}

As shown in Table~\ref{tab:dad_ablation}, DAD differs fundamentally from the decoding strategy in~\cite{xu2025wamDIFF}. Their method enforces a fixed causal or reverse-causal decoding order throughout inference, whereas DAD estimates the confidence distribution across trajectory tokens as a proxy for driving risk and dynamically selects the appropriate decoding direction in continuous scenes. This adaptive mechanism enables the model to switch between near-to-far and far-to-near inference according to scene conditions. Consequently, DAD achieves a PDMS of 88.9, outperforming R-Causal and Causal decoding by 2.4 and 7.5 points, respectively, demonstrating the effectiveness of dynamically adjusting the decoding order. The upper example in Fig.~\ref{fig:two-subfigures}(a) shows a low-risk free-driving scenario toward the front-right, where $\Phi_{prox}=0.951\pm0.014>\Phi_{dist}=0.049\pm0.007$, leading DAD to adopt Near-to-Far decoding. In contrast, the lower example is a high-risk left-turn car-following scenario, where the model conservatively determines the trajectory endpoint first, with $\Phi_{dist}=0.948\pm0.017>\Phi_{prox}=0.052\pm0.009$, resulting in Far-to-Near decoding. Notably, DAD enables frame-level confidence-guided adaptation, whereas fixed decoding strategies cannot dynamically switch modes across consecutive frames.

As shown in Table~\ref{tab:discretization_sensitivity}, trajectory discretization substantially improves driving performance over the continuous baseline. Among the tested granularities, $\Delta g=0.25$ achieves the highest PDMS of 81.4, indicating the best trade-off between quantization precision and vocabulary size.

\section{Conclusion}

We identify key training and inference challenges of discrete diffusion for autonomous driving and propose a multi-stage optimization pipeline for standard VLM architectures. Experiments on closed- and open-loop benchmarks validate its effectiveness, demonstrating the potential of discrete diffusion for efficient driving generation.


\bibliography{reference}

@misc{nvidia2026alpamayor1bridgingreasoningaction,
      title={Alpamayo-R1: Bridging Reasoning and Action Prediction for Generalizable Autonomous Driving in the Long Tail}, 
      author={NVIDIA},
      year={2026},
      eprint={2511.00088},
      archivePrefix={arXiv},
      primaryClass={cs.RO},
      url={https://arxiv.org/abs/2511.00088}, 
}

@misc{openai2024gpt4technicalreport,
      title={GPT-4 Technical Report}, 
      author={OpenAI},
      year={2024},
      eprint={2303.08774},
      archivePrefix={arXiv},
      primaryClass={cs.CL},
      url={https://arxiv.org/abs/2303.08774}, 
}

@InProceedings{The_Pitfalls_of_Next-Token_Prediction,
  title = 	 {The Pitfalls of Next-Token Prediction},
  author =       {Bachmann, Gregor and Nagarajan, Vaishnavh},
  booktitle = 	 {Proceedings of the 41st International Conference on Machine Learning},
  pages = 	 {2296--2318},
  year = 	 {2024},
  editor = 	 {Salakhutdinov, Ruslan and Kolter, Zico and Heller, Katherine and Weller, Adrian and Oliver, Nuria and Scarlett, Jonathan and Berkenkamp, Felix},
  volume = 	 {235},
  series = 	 {Proceedings of Machine Learning Research},
  month = 	 {21--27 Jul},
  publisher =    {PMLR},
  url = 	 {https://proceedings.mlr.press/v235/bachmann24a.html}
}

@InProceedings{Computational-Statistical-Tradeoffs-at-the-Next-Token-Prediction,
  title = 	 {Computational-Statistical Tradeoffs at the Next-Token Prediction Barrier: Autoregressive and Imitation Learning under Misspecification (extended abstract)},
  author =       {Rohatgi, Dhruv and Block, Adam and Huang, Audrey and Krishnamurthy, Akshay and Foster, Dylan J.},
  booktitle = 	 {Proceedings of Thirty Eighth Conference on Learning Theory},
  pages = 	 {4831--4837},
  year = 	 {2025},
  editor = 	 {Haghtalab, Nika and Moitra, Ankur},
  volume = 	 {291},
  series = 	 {Proceedings of Machine Learning Research},
  month = 	 {30 Jun--04 Jul},
  publisher =    {PMLR},
  url = 	 {https://proceedings.mlr.press/v291/rohatgi25a.html},
}

@misc{labs2025mercuryultrafastlanguagemodels,
      title={Mercury: Ultra-Fast Language Models Based on Diffusion}, 
      author={Inception Labs and Samar Khanna and Siddhant Kharbanda and Shufan Li and Harshit Varma and Eric Wang and Sawyer Birnbaum and Ziyang Luo and Yanis Miraoui and Akash Palrecha and Stefano Ermon and Aditya Grover and Volodymyr Kuleshov},
      year={2025},
      eprint={2506.17298},
      archivePrefix={arXiv},
      primaryClass={cs.CL},
      url={https://arxiv.org/abs/2506.17298}, 
}

@misc{geminidiffusion,
  author = {{DeepMind}},
  title = {Gemini Diffusion},
  year = {2025},
  howpublished = {\url{https://deepmind.google/models/gemini-diffusion/}},
}

@misc{gong2025scalingdiffusionlanguagemodels,
      title={Scaling Diffusion Language Models via Adaptation from Autoregressive Models}, 
      author={Shansan Gong and Shivam Agarwal and Yizhe Zhang and Jiacheng Ye and Lin Zheng and Mukai Li and Chenxin An and Peilin Zhao and Wei Bi and Jiawei Han and Hao Peng and Lingpeng Kong},
      year={2025},
      eprint={2410.17891},
      archivePrefix={arXiv},
      primaryClass={cs.CL},
      url={https://arxiv.org/abs/2410.17891}, 
}

@misc{gong2025diffucoderunderstandingimprovingmasked,
      title={DiffuCoder: Understanding and Improving Masked Diffusion Models for Code Generation}, 
      author={Shansan Gong and Ruixiang Zhang and Huangjie Zheng and Jiatao Gu and Navdeep Jaitly and Lingpeng Kong and Yizhe Zhang},
      year={2025},
      eprint={2506.20639},
      archivePrefix={arXiv},
      primaryClass={cs.CL},
      url={https://arxiv.org/abs/2506.20639}, 
}

@inproceedings{
Large-Language-Diffusion-Models,
title={Large Language Diffusion Models},
author={Shen Nie and Fengqi Zhu and Zebin You and Xiaolu Zhang and Jingyang Ou and Jun Hu and JUN ZHOU and Yankai Lin and Ji-Rong Wen and Chongxuan Li},
booktitle={The Thirty-ninth Annual Conference on Neural Information Processing Systems},
year={2025},
url={https://openreview.net/forum?id=KnqiC0znVF}
}

@misc{d1,
      title={d1: Scaling Reasoning in Diffusion Large Language Models via Reinforcement Learning}, 
      author={Siyan Zhao and Devaansh Gupta and Qinqing Zheng and Aditya Grover},
      year={2025},
      eprint={2504.12216},
      archivePrefix={arXiv},
      primaryClass={cs.CL},
      url={https://arxiv.org/abs/2504.12216}, 
}

@misc{wang2024qwen2vlenhancingvisionlanguagemodels,
      title={Qwen2-VL: Enhancing Vision-Language Model's Perception of the World at Any Resolution}, 
      author={Peng Wang and Shuai Bai and Sinan Tan and Shijie Wang and Zhihao Fan and Jinze Bai and Keqin Chen and Xuejing Liu and Jialin Wang and Wenbin Ge and Yang Fan and Kai Dang and Mengfei Du and Xuancheng Ren and Rui Men and Dayiheng Liu and Chang Zhou and Jingren Zhou and Junyang Lin},
      year={2024},
      eprint={2409.12191},
      archivePrefix={arXiv},
      primaryClass={cs.CV},
      url={https://arxiv.org/abs/2409.12191}, 
}

@misc{bai2025qwen25vltechnicalreport,
      title={Qwen2.5-VL Technical Report}, 
      author={Shuai Bai and Keqin Chen and Xuejing Liu and Jialin Wang and Wenbin Ge and Sibo Song and Kai Dang and Peng Wang and Shijie Wang and Jun Tang and Humen Zhong and Yuanzhi Zhu and Mingkun Yang and Zhaohai Li and Jianqiang Wan and Pengfei Wang and Wei Ding and Zheren Fu and Yiheng Xu and Jiabo Ye and Xi Zhang and Tianbao Xie and Zesen Cheng and Hang Zhang and Zhibo Yang and Haiyang Xu and Junyang Lin},
      year={2025},
      eprint={2502.13923},
      archivePrefix={arXiv},
      primaryClass={cs.CV},
      url={https://arxiv.org/abs/2502.13923}, 
}

@misc{yang2025qwen3technicalreport,
      title={Qwen3 Technical Report}, 
      author={An Yang and Anfeng Li and Baosong Yang and Beichen Zhang and Binyuan Hui and Bo Zheng and Bowen Yu and Chang Gao and Chengen Huang and Chenxu Lv and Chujie Zheng and Dayiheng Liu and Fan Zhou and Fei Huang and Feng Hu and Hao Ge and Haoran Wei and Huan Lin and Jialong Tang and Jian Yang and Jianhong Tu and Jianwei Zhang and Jianxin Yang and Jiaxi Yang and Jing Zhou and Jingren Zhou and Junyang Lin and Kai Dang and Keqin Bao and Kexin Yang and Le Yu and Lianghao Deng and Mei Li and Mingfeng Xue and Mingze Li and Pei Zhang and Peng Wang and Qin Zhu and Rui Men and Ruize Gao and Shixuan Liu and Shuang Luo and Tianhao Li and Tianyi Tang and Wenbiao Yin and Xingzhang Ren and Xinyu Wang and Xinyu Zhang and Xuancheng Ren and Yang Fan and Yang Su and Yichang Zhang and Yinger Zhang and Yu Wan and Yuqiong Liu and Zekun Wang and Zeyu Cui and Zhenru Zhang and Zhipeng Zhou and Zihan Qiu},
      year={2025},
      eprint={2505.09388},
      archivePrefix={arXiv},
      primaryClass={cs.CL},
      url={https://arxiv.org/abs/2505.09388}, 
}

@inproceedings{guo-etal-2025-mammoth,
    title = "{MA}mmo{TH}-{VL}: Eliciting Multimodal Reasoning with Instruction Tuning at Scale",
    author = "Guo, Jiawei  and
      Zheng, Tianyu  and
      Li, Yizhi  and
      Bai, Yuelin  and
      Li, Bo  and
      Wang, Yubo  and
      Zhu, King  and
      Neubig, Graham  and
      Chen, Wenhu  and
      Yue, Xiang",
    editor = "Che, Wanxiang  and
      Nabende, Joyce  and
      Shutova, Ekaterina  and
      Pilehvar, Mohammad Taher",
    booktitle = "Proceedings of the 63rd Annual Meeting of the Association for Computational Linguistics (Volume 1: Long Papers)",
    month = jul,
    year = "2025",
    address = "Vienna, Austria",
    publisher = "Association for Computational Linguistics",
    url = "https://aclanthology.org/2025.acl-long.680/",
    doi = "10.18653/v1/2025.acl-long.680",
    pages = "13869--13920",
    ISBN = "979-8-89176-251-0"
}

@inproceedings{
li2025lavida,
title={LaViDa: A Large Diffusion Model for Vision-Language Understanding},
author={Shufan Li and Konstantinos Kallidromitis and Hritik Bansal and Akash Gokul and Yusuke Kato and Kazuki Kozuka and Jason Kuen and Zhe Lin and Kai-Wei Chang and Aditya Grover},
booktitle={The Thirty-ninth Annual Conference on Neural Information Processing Systems},
year={2025},
url={https://openreview.net/forum?id=6WnBITpnzD}
}

@misc{you2025lladavlargelanguagediffusion,
      title={LLaDA-V: Large Language Diffusion Models with Visual Instruction Tuning}, 
      author={Zebin You and Shen Nie and Xiaolu Zhang and Jun Hu and Jun Zhou and Zhiwu Lu and Ji-Rong Wen and Chongxuan Li},
      year={2025},
      eprint={2505.16933},
      archivePrefix={arXiv},
      primaryClass={cs.LG},
      url={https://arxiv.org/abs/2505.16933}, 
}

@inproceedings{
MMaDA,
title={Parallel Multimodal Diffusion Language Models for Thinking-Aware Editing and Generation},
author={Ye Tian and Ling Yang and JiongFan Yang and Anran Wang and Yu Tian and Jiani zheng and Haochen Wang and Zhiyang Teng and Zhuochen Wang and Yinjie Wang and Yunhai Tong and Mengdi Wang and Xiangtai Li},
booktitle={The Fourteenth International Conference on Learning Representations},
year={2026},
url={https://openreview.net/forum?id=mkQAd11ovn}
}

@misc{yu2025dimplediscretediffusionmultimodal,
      title={Dimple: Discrete Diffusion Multimodal Large Language Model with Parallel Decoding}, 
      author={Runpeng Yu and Xinyin Ma and Xinchao Wang},
      year={2025},
      eprint={2505.16990},
      archivePrefix={arXiv},
      primaryClass={cs.CV},
      url={https://arxiv.org/abs/2505.16990}, 
}

@misc{li2025lavidaoelasticlargemasked,
      title={Lavida-O: Elastic Large Masked Diffusion Models for Unified Multimodal Understanding and Generation}, 
      author={Shufan Li and Jiuxiang Gu and Kangning Liu and Zhe Lin and Zijun Wei and Aditya Grover and Jason Kuen},
      year={2025},
      eprint={2509.19244},
      archivePrefix={arXiv},
      primaryClass={cs.CV},
      url={https://arxiv.org/abs/2509.19244}, 
}

@misc{ye2026dreamvldreamvlaopen,
      title={Dream-VL and Dream-VLA: Open Vision-Language and Vision-Language-Action Models with Diffusion Language Model Backbone}, 
      author={Jiacheng Ye and Shansan Gong and Jiahui Gao and Junming Fan and Shuang Wu and Wei Bi and Haoli Bai and Lifeng Shang and Lingpeng Kong},
      year={2026},
      eprint={2512.22615},
      archivePrefix={arXiv},
      primaryClass={cs.CV},
      url={https://arxiv.org/abs/2512.22615}, 
}

@misc{cui2025viladlargevisionlanguage,
      title={ViLaD: A Large Vision Language Diffusion Framework for End-to-End Autonomous Driving}, 
      author={Can Cui and Yupeng Zhou and Juntong Peng and Sung-Yeon Park and Zichong Yang and Prashanth Sankaranarayanan and Jiaru Zhang and Ruqi Zhang and Ziran Wang},
      year={2025},
      eprint={2508.12603},
      archivePrefix={arXiv},
      primaryClass={cs.CV},
      url={https://arxiv.org/abs/2508.12603}, 
}

@misc{zhang2026efficientexplainableendtoendautonomous,
      title={Efficient and Explainable End-to-End Autonomous Driving via Masked Vision-Language-Action Diffusion}, 
      author={Jiaru Zhang and Manav Gagvani and Can Cui and Juntong Peng and Ruqi Zhang and Ziran Wang},
      year={2026},
      eprint={2602.20577},
      archivePrefix={arXiv},
      primaryClass={cs.CV},
      url={https://arxiv.org/abs/2602.20577}, 
}

@InProceedings{Zero-Shot-Text-to-Image-Generation,
  title = 	 {Zero-Shot Text-to-Image Generation},
  author =       {Ramesh, Aditya and Pavlov, Mikhail and Goh, Gabriel and Gray, Scott and Voss, Chelsea and Radford, Alec and Chen, Mark and Sutskever, Ilya},
  booktitle = 	 {Proceedings of the 38th International Conference on Machine Learning},
  pages = 	 {8821--8831},
  year = 	 {2021},
  editor = 	 {Meila, Marina and Zhang, Tong},
  volume = 	 {139},
  series = 	 {Proceedings of Machine Learning Research},
  month = 	 {18--24 Jul},
  publisher =    {PMLR},
  url = 	 {https://proceedings.mlr.press/v139/ramesh21a.html}
}

@article{
Scaling-Autoregressive-Models-for-Content-Rich-Text-to-Image-Generation,
title={Scaling Autoregressive Models for Content-Rich Text-to-Image Generation},
author={Jiahui Yu and Yuanzhong Xu and Jing Yu Koh and Thang Luong and Gunjan Baid and Zirui Wang and Vijay Vasudevan and Alexander Ku and Yinfei Yang and Burcu Karagol Ayan and Ben Hutchinson and Wei Han and Zarana Parekh and Xin Li and Han Zhang and Jason Baldridge and Yonghui Wu},
journal={Transactions on Machine Learning Research},
issn={2835-8856},
year={2022},
url={https://openreview.net/forum?id=AFDcYJKhND},
note={Featured Certification}
}

@misc{oord2016wavenetgenerativemodelraw,
      title={WaveNet: A Generative Model for Raw Audio}, 
      author={Aaron van den Oord and Sander Dieleman and Heiga Zen and Karen Simonyan and Oriol Vinyals and Alex Graves and Nal Kalchbrenner and Andrew Senior and Koray Kavukcuoglu},
      year={2016},
      eprint={1609.03499},
      archivePrefix={arXiv},
      primaryClass={cs.SD},
      url={https://arxiv.org/abs/1609.03499}, 
}

@misc{seff2023motionlmmultiagentmotionforecasting,
      title={MotionLM: Multi-Agent Motion Forecasting as Language Modeling}, 
      author={Ari Seff and Brian Cera and Dian Chen and Mason Ng and Aurick Zhou and Nigamaa Nayakanti and Khaled S. Refaat and Rami Al-Rfou and Benjamin Sapp},
      year={2023},
      eprint={2309.16534},
      archivePrefix={arXiv},
      primaryClass={cs.CV},
      url={https://arxiv.org/abs/2309.16534}, 
}

@inproceedings{Structured-denoising-diffusion-models-in-discrete-state-spaces,
author = {Austin, Jacob and Johnson, Daniel D. and Ho, Jonathan and Tarlow, Daniel and van den Berg, Rianne},
title = {Structured denoising diffusion models in discrete state-spaces},
year = {2021},
isbn = {9781713845393},
publisher = {Curran Associates Inc.},
address = {Red Hook, NY, USA},
booktitle = {Proceedings of the 35th International Conference on Neural Information Processing Systems},
articleno = {1376},
numpages = {13},
series = {NIPS '21}
}

@misc{zhang2026onpolicyrlmeetsoffpolicy,
      title={On-Policy RL Meets Off-Policy Experts: Harmonizing Supervised Fine-Tuning and Reinforcement Learning via Dynamic Weighting}, 
      author={Wenhao Zhang and Yuexiang Xie and Yuchang Sun and Yanxi Chen and Guoyin Wang and Yaliang Li and Bolin Ding and Jingren Zhou},
      year={2026},
      eprint={2508.11408},
      archivePrefix={arXiv},
      primaryClass={cs.LG},
      url={https://arxiv.org/abs/2508.11408}, 
}

@misc{shao2024deepseekmathpushinglimitsmathematical,
      title={DeepSeekMath: Pushing the Limits of Mathematical Reasoning in Open Language Models}, 
      author={Zhihong Shao and Peiyi Wang and Qihao Zhu and Runxin Xu and Junxiao Song and Xiao Bi and Haowei Zhang and Mingchuan Zhang and Y. K. Li and Y. Wu and Daya Guo},
      year={2024},
      eprint={2402.03300},
      archivePrefix={arXiv},
      primaryClass={cs.CL},
      url={https://arxiv.org/abs/2402.03300}, 
}

@misc{zhang2020mapadaptivegoalbasedtrajectoryprediction,
      title={Map-Adaptive Goal-Based Trajectory Prediction}, 
      author={Lingyao Zhang and Po-Hsun Su and Jerrick Hoang and Galen Clark Haynes and Micol Marchetti-Bowick},
      year={2020},
      eprint={2009.04450},
      archivePrefix={arXiv},
      primaryClass={cs.LG},
      url={https://arxiv.org/abs/2009.04450}, 
}

@INPROCEEDINGS {nuScenes,
author = { Caesar, Holger and Bankiti, Varun and Lang, Alex H. and Vora, Sourabh and Liong, Venice Erin and Xu, Qiang and Krishnan, Anush and Pan, Yu and Baldan, Giancarlo and Beijbom, Oscar },
booktitle = { 2020 IEEE/CVF Conference on Computer Vision and Pattern Recognition (CVPR) },
title = {{ nuScenes: A Multimodal Dataset for Autonomous Driving }},
year = {2020},
volume = {},
ISSN = {},
pages = {11618-11628},
doi = {10.1109/CVPR42600.2020.01164},
url = {https://doi.ieeecomputersociety.org/10.1109/CVPR42600.2020.01164},
publisher = {IEEE Computer Society},
address = {Los Alamitos, CA, USA},
month =Jun}

@misc{wang2024drivecot,
      title={DriveCoT: Integrating Chain-of-Thought Reasoning with End-to-End Driving}, 
      author={Tianqi Wang and Enze Xie and Ruihang Chu and Zhenguo Li and Ping Luo},
      year={2024},
      eprint={2403.16996},
      archivePrefix={arXiv},
      primaryClass={cs.CV},
      url={https://arxiv.org/abs/2403.16996}, 
}

@inproceedings{zhang2025epona,
  author = {Zhang, Kaiwen and Tang, Zhenyu and Hu, Xiaotao and Pan, Xingang and Guo, Xiaoyang and Liu, Yuan and Huang,
  Jingwei and Yuan, Li and Zhang, Qian and Long, Xiao-Xiao and Cao, Xun and Yin, Wei},
  title = {Epona: Autoregressive Diffusion World Model for Autonomous Driving},
  booktitle = {Proceedings of the IEEE/CVF International Conference on Computer Vision (ICCV)},
  year = {2025}
}

@misc{chen2024drivinggptunifyingdrivingworld,
      title={DrivingGPT: Unifying Driving World Modeling and Planning with Multi-modal Autoregressive Transformers}, 
      author={Yuntao Chen and Yuqi Wang and Zhaoxiang Zhang},
      year={2024},
      eprint={2412.18607},
      archivePrefix={arXiv},
      primaryClass={cs.CV},
      url={https://arxiv.org/abs/2412.18607}, 
}

@inproceedings{
zeng2025futuresightdrive,
title={FutureSightDrive: Thinking Visually with Spatio-Temporal CoT for Autonomous Driving},
author={Shuang Zeng and Xinyuan Chang and Mengwei Xie and Xinran Liu and Yifan Bai and Zheng Pan and Mu Xu and Xing Wei},
booktitle={The Thirty-ninth Annual Conference on Neural Information Processing Systems},
year={2025},
url={https://openreview.net/forum?id=fCirUh6FRb}
}

@misc{li2025recogdrive,
      title={ReCogDrive: A Reinforced Cognitive Framework for End-to-End Autonomous Driving}, 
      author={Yongkang Li and Kaixin Xiong and Xiangyu Guo and Fang Li and Sixu Yan and Gangwei Xu and Lijun Zhou and Long Chen and Haiyang Sun and Bing Wang and Kun Ma and Guang Chen and Hangjun Ye and Wenyu Liu and Xinggang Wang},
      year={2025},
      eprint={2506.08052},
      archivePrefix={arXiv},
      primaryClass={cs.CV},
      url={https://arxiv.org/abs/2506.08052}, 
}

@article{TransFuser,
author = {Chitta, Kashyap and Prakash, Aditya and Jaeger, Bernhard and Yu, Zehao and Renz, Katrin and Geiger, Andreas},
title = {TransFuser: Imitation With Transformer-Based Sensor Fusion for Autonomous Driving},
year = {2023},
issue_date = {Nov. 2023},
publisher = {IEEE Computer Society},
address = {USA},
volume = {45},
number = {11},
issn = {0162-8828},
url = {https://doi.org/10.1109/TPAMI.2022.3200245},
doi = {10.1109/TPAMI.2022.3200245},
journal = {IEEE Trans. Pattern Anal. Mach. Intell.},
month = nov,
pages = {12878–12895},
numpages = {18}
}

@inproceedings{hu2023_uniad,
 title={Planning-oriented Autonomous Driving}, 
 author={Yihan Hu and Jiazhi Yang and Li Chen and Keyu Li and Chonghao Sima and Xizhou Zhu and Siqi Chai and Senyao Du and Tianwei Lin and Wenhai Wang and Lewei Lu and Xiaosong Jia and Qiang Liu and Jifeng Dai and Yu Qiao and Hongyang Li},
 booktitle={Proceedings of the IEEE/CVF Conference on Computer Vision and Pattern Recognition},
 year={2023},
}

@inproceedings{
jiang2026vadv2,
title={{VAD}v2: End-to-End Autonomous Driving via Probabilistic Planning},
author={Bo Jiang and Shaoyu Chen and Hao Gao and Bencheng Liao and Qian Zhang and Wenyu Liu and Xinggang Wang},
booktitle={The Fourteenth International Conference on Learning Representations},
year={2026},
url={https://openreview.net/forum?id=0a4dA6eUHN}
}

@INPROCEEDINGS{PARA-Drive,
  author={Weng, Xinshuo and Ivanovic, Boris and Wang, Yan and Wang, Yue and Pavone, Marco},
  booktitle={2024 IEEE/CVF Conference on Computer Vision and Pattern Recognition (CVPR)}, 
  title={PARA-Drive: Parallelized Architecture for Real-Time Autonomous Driving}, 
  year={2024},
  volume={},
  number={},
  pages={15449-15458},
  doi={10.1109/CVPR52733.2024.01463}}

@misc{li2024hydramdp,
      title={Hydra-MDP: End-to-end Multimodal Planning with Multi-target Hydra-Distillation}, 
      author={Zhenxin Li and Kailin Li and Shihao Wang and Shiyi Lan and Zhiding Yu and Yishen Ji and Zhiqi Li and Ziyue Zhu and Jan Kautz and Zuxuan Wu and Yu-Gang Jiang and Jose M. Alvarez},
      year={2024},
      eprint={2406.06978},
      archivePrefix={arXiv},
      primaryClass={cs.CV},
      url={https://arxiv.org/abs/2406.06978}, 
}

@misc{yuan2024drama,
      title={DRAMA: An Efficient End-to-end Motion Planner for Autonomous Driving with Mamba}, 
      author={Chengran Yuan and Zhanqi Zhang and Jiawei Sun and Shuo Sun and Zefan Huang and Christina Dao Wen Lee and Dongen Li and Yuhang Han and Anthony Wong and Keng Peng Tee and Marcelo H. Ang Jr},
      year={2024},
      eprint={2408.03601},
      archivePrefix={arXiv},
      primaryClass={cs.RO},
      url={https://arxiv.org/abs/2408.03601}, 
}

@misc{openemma,
      title={OpenEMMA: Open-Source Multimodal Model for End-to-End Autonomous Driving}, 
      author={Shuo Xing and Chengyuan Qian and Yuping Wang and Hongyuan Hua and Kexin Tian and Yang Zhou and Zhengzhong Tu},
      year={2025},
      eprint={2412.15208},
      archivePrefix={arXiv},
      primaryClass={cs.CV},
      url={https://arxiv.org/abs/2412.15208}, 
}

@article{li2025drivevlaW0,
  title={DriveVLA-W0: World models amplify data scaling law in autonomous driving},
  author={Li, Yingyan and Shang, Shuyao and Liu, Weisong and Zhan, Bing and Wang, Haochen and Wang, Yuqi and Chen, Yuntao and Wang, Xiaoman and An, Yasong and Tang, Chufeng and others},
  journal={arXiv preprint arXiv:2510.12796},
  year={2025}
}

@article{diffusiondrive,
  title={DiffusionDrive: Truncated Diffusion Model for End-to-End Autonomous Driving},
  author={Bencheng Liao and Shaoyu Chen and Haoran Yin and Bo Jiang and Cheng Wang and Sixu Yan and Xinbang Zhang and Xiangyu Li and Ying Zhang and Qian Zhang and Xinggang Wang},
  booktitle    = {{IEEE/CVF} Conference on Computer Vision and Pattern Recognition,
                  {CVPR} 2025, Nashville, TN, USA, June 11-15, 2025},
  pages        = {12037--12047},
  publisher    = {Computer Vision Foundation / {IEEE}},
  year         = {2025},
  url          = {https://openaccess.thecvf.com/content/CVPR2025/html/Liao\_DiffusionDrive\_Truncated\_Diffusion\_Model\_for\_End-to-End\_Autonomous\_Driving\_CVPR\_2025\_paper.html},
  doi          = {10.1109/CVPR52734.2025.01124}
}

@article{xu2025wamFLOW,
  title={WAM-Flow: Parallel Coarse-to-Fine Motion Planning via Discrete Flow Matching for Autonomous Driving},
  author={Xu, Yifang and Cui, Jiahao and Cai, Feipeng and Zhu, Zhihao and Shang, Hanlin and Luan, Shan and Xu, Mingwang and Zhang, Neng and Li, Yaoyi and Cai, Jia and others},
  journal={arXiv preprint arXiv:2512.06112},
  year={2025}
}

@article{xu2025wamDIFF,
  title={WAM-Diff: A Masked Diffusion VLA Framework with MoE and Online Reinforcement Learning for Autonomous Driving},
  author={Xu, Mingwang and Cui, Jiahao and Cai, Feipeng and Shang, Hanlin and Zhu, Zhihao and Luan, Shan and Xu, Yifang and Zhang, Neng and Li, Yaoyi and Cai, Jia and others},
  journal={arXiv preprint arXiv:2512.11872},
  year={2025}
}


\end{document}